\documentclass{article}

\usepackage[preprint]{tmlr}

\usepackage[utf8]{inputenc} 
\usepackage[T1]{fontenc}    
\usepackage{hyperref}       
\usepackage{url}            
\usepackage{booktabs}       
\usepackage{amsfonts}       
\usepackage{nicefrac}       
\usepackage{microtype}      
\usepackage{xcolor}         
\usepackage{dsfont}         

\usepackage{amsmath,amsfonts,bm}

\def\eqref#1{equation~\ref{#1}}

\def\1{\bm{1}}

\DeclareMathAlphabet{\mathsfit}{\encodingdefault}{\sfdefault}{m}{sl}
\SetMathAlphabet{\mathsfit}{bold}{\encodingdefault}{\sfdefault}{bx}{n}

\newcommand{\KL}{D_{\mathrm{KL}}}

\DeclareMathOperator*{\argmax}{arg\,max}

\newcommand{\bone}{\mathbf{1}}

\usepackage{url}
\usepackage{booktabs}
\usepackage{amsmath,amsthm,amsfonts,amssymb,amscd}
\usepackage{bbm}
\usepackage{cleveref}
\usepackage{lastpage}
\usepackage{enumerate}
\usepackage{fancyhdr}
\usepackage{mathrsfs}
\usepackage{xcolor}
\usepackage{graphicx}
\usepackage{listings}
\usepackage{mathtools}
\usepackage{todonotes}
\usepackage{adjustbox}
\usepackage{wrapfig}

\usepackage{algorithm,algpseudocode}
\usepackage{natbib}
\usepackage{multirow}
\usepackage{array}
\usepackage{dblfloatfix}
\usepackage{makecell}
\usepackage{spverbatim}
\usepackage{appendix}
\usepackage{bbm}
\usepackage{comment}
\usepackage{subcaption}
\usepackage{alltt}

\newcommand{\expect}{\ensuremath{\mathop{\mathbb{E}}}}

\usepackage{enumitem}

\newcommand{\experiment}{\xi}

\usepackage{titletoc}

\title{Induction and Inquiry via\\ Probabilistic Reasoning over Language and Code}

\author{Wasu Top Piriyakulkij$^{*1}$ Sam Acquaviva$^{*2}$ Cassidy Langenfeld$^{1}$ Joshua Tenenbaum$^2$ Kevin Ellis$^{1}$\\
$^1$Cornell University$\quad$ $^2$Massachusetts Institute of Technology$\quad$ $^*$Equal Contribution\\
Correspondence to: \texttt{wp237@cornell.edu}, \texttt{samacqua@gmail.com}, \texttt{kellis@cornell.edu}
}

\begin{document}
\captionsetup[subfigure]{font={small}, skip=0pt, singlelinecheck=false}

\maketitle

\begin{abstract}
How humans grow and maintain abstract knowledge from the sparse, streaming noisy data of experience is a longstanding challenge in cognitive science.
Any computational account must satisfy at least three desiderata: It must be (1) data-efficient and compute-efficient, (2) capture gradations of uncertainty to support intelligent inquiry and information gathering, and (3) be flexible enough to mentally represent the endless range of concepts people can learn and think about.
Here we introduce a computational model that captures these three properties, by encoding symbolic knowledge as mental programs that combine natural language with source code, and
 sequentially inferring mental programs using LLM-guided Bayesian learning algorithms.
Across a range of behavioral studies this model successfully reproduces quantitative signatures of human inductive learning and active inquiry, such as anchoring, garden-pathing, and other effects.
In contrast, pure LLMs and classic Bayesian models either fail at the underlying task, or do not reproduce human behavior, or succeed only at exorbitant computational cost.
These results suggest that one way humans continually grow their knowledge is by mentally representing many hypotheses spanning language-like and program-like representations, then revising those hypotheses to approximate Bayesian updates, while a bottom-up neural mechanism (an LLM) makes inference both tractable and learnable.
\end{abstract}

\section{Introduction}

Inductive reasoning is a cornerstone of general intelligence:
Learning new concepts from few examples, 
and revising those concepts in light of new evidence.
Limited data is inherently ambiguous, motivating an inquiry process of asking questions or doing experiments to resolve uncertainty.
This induction-inquiry cycle unfolds sequentially over time, with new data streaming in, because inquiry is an active process of asking questions and getting answers.
Modeling human induction and inquiry is a longstanding challenge because such models must handle uncertainty, have a flexible hypothesis class covering much of what humans can think of, and support efficient online computation.
These objectives interact:
A flexible, open-ended hypothesis class yields more uncertainty, because there are now more competing explanations for the evidence.
But this causes reasoning to be 
computationally expensive.
Decades of research~\cite{anderson1990theac,https://doi.org/10.1111/tops.12142,griffiths2024bayesian} suggest human inductive reasoning approximates probabilistic Bayesian belief updates, but
we still cannot truly model what people seem to do:
Efficient online induction and inquiry over flexible open-ended hypothesis spaces.
This is the challenge we take on.

We start with the Bayesian cognitive modeling paradigm, which imposes probabilistic norms for calculating how credible a belief should be, but as a paradigm, does not say what people can believe in the first place---how they can efficiently reason about an endlessly open-ended range of concepts.
Prior models of inductive reasoning~\cite{quilty2023best,piantadosi2011learning} further posit an inner \emph{Language of Thought}, whether formal logic, symbolic schemas or Bayes net templates, or probabilistic programs, which  formalize and delineate what hypotheses are representable, and therefore learnable.
The literature on intuitive theories and cognitive development has also proposed  natural language as a representation of hypotheses~\cite{carey2011origin,whatmakesussmart,spelke2022babies}, but this has never been made formal.



\begin{figure*}[t!]
\centering

\includegraphics[width=\linewidth]{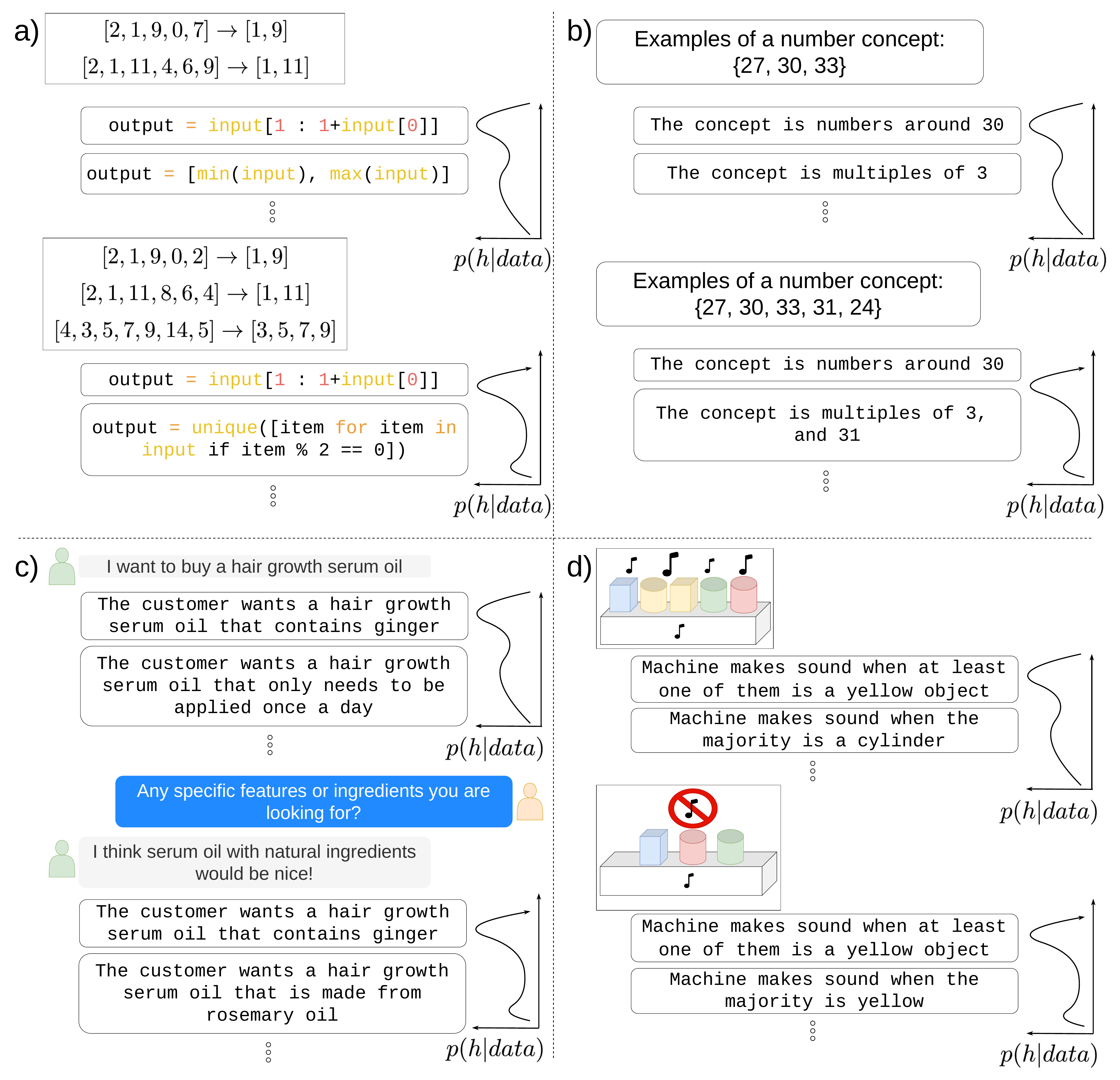}
\vspace{-1em}
\caption{a) - d) show sequential inference problems that we study in this work and illustrate how, in each problem, Bayesian beliefs may change upon seeing new observations over time.}
\label{fig:domain}
\vspace{-1em}
\end{figure*}

Here we find that human behavior across a range of induction and inquiry setups is best explained by sequential probabilistic reasoning over 
\emph{mental programs}, which we treat as a mix of natural language and computer source code (\cref{fig:domain}).
Although the idea of an inner Language of Thought is an old one, its past computational instantiations assumed rigid logical forms that are less malleable than natural language, and less practical than actual programming languages.

Why represent knowledge as a mix of natural language and source code?
Language and code are generic representations for communicating and formalizing human knowledge, but
only recently have they become tractable targets of inference, owing primarily to Large Language Models (LLMs).
Our models equip LLMs with sequential probabilistic reasoning. 
The resulting models
reproduce sequential phenomena such as garden-pathing and anchoring; capture gradations of uncertainty; and scale to more complex concepts, because of the powerful combination of the expressivity of language and the top-down feedback of code.
Furthermore, we show how these models can perform human-like active inquiry, closing the sequential learning loop which alternates between induction and inquiry.
Speculatively, 
our model suggests mental representations that lie on a continuum between logic and language, and shows how this representation works with Bayesian reasoning.

\begin{figure*}[t!]
\centering

\includegraphics[width=\linewidth]{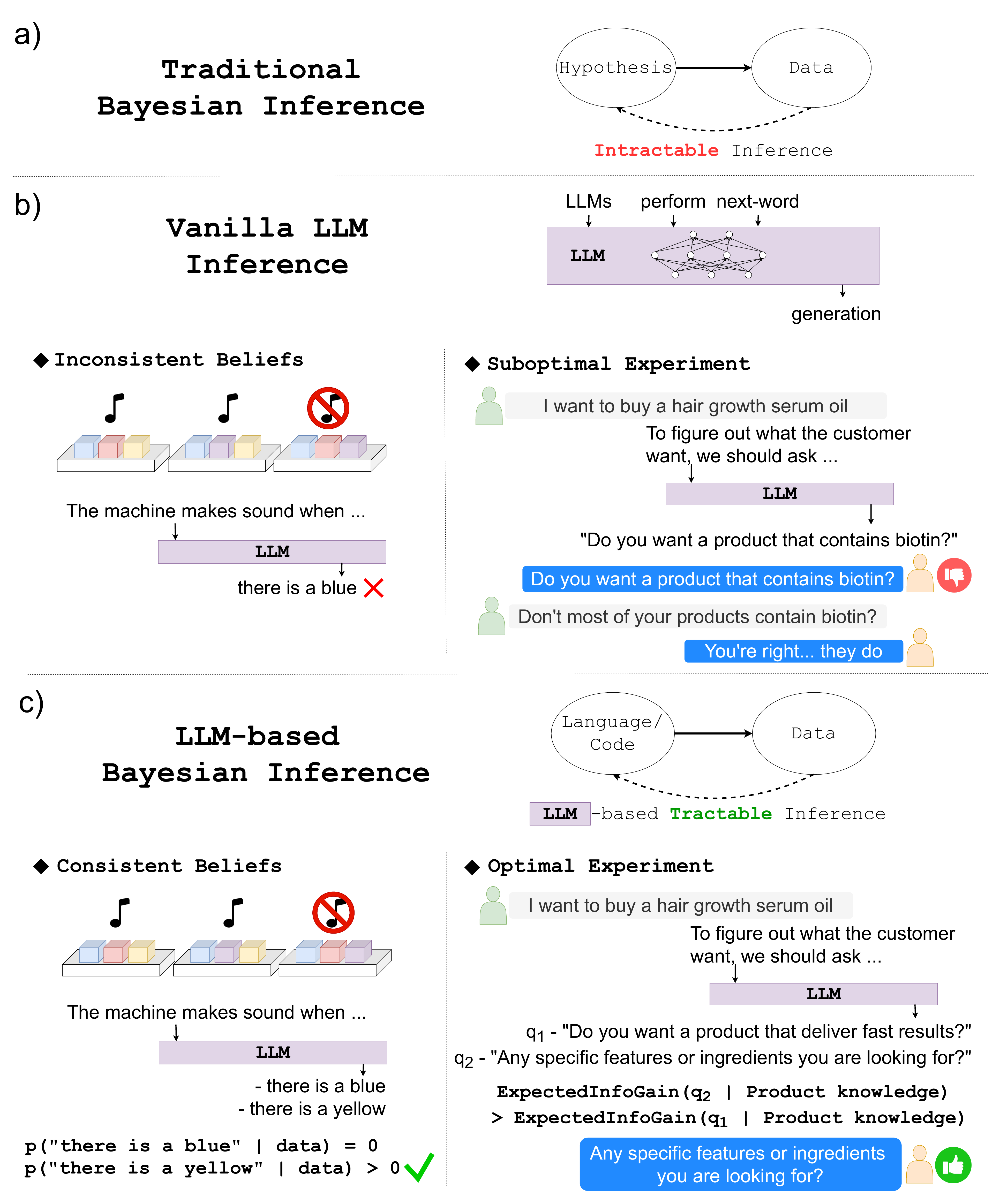}
\vspace{-1em}
\caption{a) - c) show three types of inference methods: traditional Bayesian, vanilla LLM, and LLM-based Bayesian (ours). LLM-based Bayesian inference in language/code hypothesis space is the only method that is tractable while maintaining consistent beliefs and optimal experiments.}
\label{fig:high_level_method}
\vspace{-1em}
\end{figure*}


\begin{figure*}[!b]
\centering

\includegraphics[width=1\linewidth]{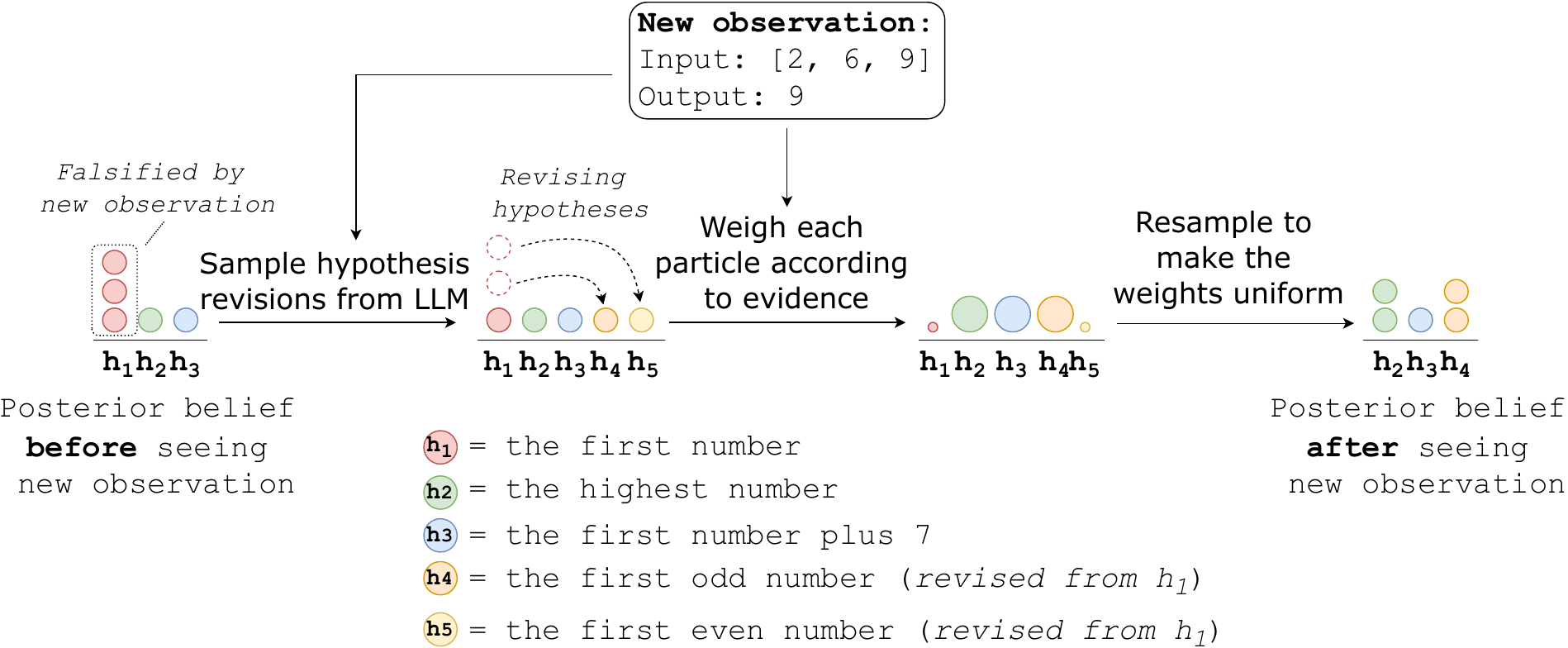}

\caption{An illustration of how Sequential Monte Carlo methods change posterior belief upon receiving new observation. Sequential Monte Carlo method tracks a small number of hypotheses (called particles) represented above by circles.
After each experiment, some particles are revised in light of the new observation, with the help of LLM. Then, the particles are reweighed according to how well each explains the observations we have seen so far. Resampling adjusts the weights of particles to be uniform by pruning low-probability hypotheses and multiplying high-probability ones.
}
\label{fig:smc}
\end{figure*}

\section{Computational Model}


Humans encounter evidence sequentially over time:
One instance of a new category is seen first, another second, etc.
Limited data is inherently ambiguous, so 
we model humans as mentally representing multiple competing hypotheses, 
maintaining those 
that both fit the data and admit simple natural-language description.
Upon receiving new evidence, humans update their beliefs:
They inductively reason about whether new data forces new conclusions, or eliminates old hypotheses.
Therefore our model compares the latest hypotheses to the data, and stochastically revises them to better fit the data.
Representing hypothesis in language and code, and then revising hypotheses using large language models, allows efficient open-ended reasoning.
Modeling multiple competing hypotheses captures the intuition that people can think of several different explanations, which allows rational inquiry by asking questions that optimally split the competing hypotheses.

Formally, given a sequence of $T$ examples $e_{1:T}$, our model hypothesizes mental programs $h$.
Each mental program has two pieces:
(1) a natural language description and (2) a Python implementation.
Mixing language and code allows freely generating ideas in natural language, but forces formalizing hypothesis into executable form.
We define priors $p(h)$ that favor short natural language descriptions, and likelihoods $p(e_t\mid h)$ that favor program executions that match the evidence.
The language prior and code likelihood together define
a posterior $p(h\mid e_{1:T})$, which evolves over time:
\begin{align}
p(h\mid e_{1:T})\propto p(e_T\mid h) p(h\mid e_{1:T-1}) \propto p(h)\prod_{t\leq T} p(e_{t}\mid h) \label{eq:beliefupdate}
\end{align}

The above posterior is intractable because infinitely many hypotheses could explain the data.
Instead, humans could only plausibly consider a small finite set of hypotheses.

How should we generate this small pool of possible hypotheses, given the vast hypothesis space of natural language and code?
While we can compare competing hypotheses given the prior and likelihood (\cref{eq:beliefupdate}), we still need a heuristic proposal mechanism to know which hypotheses to consider in the first place.
LLMs are a natural choice.
From a cognitive perspective, they are a fast bottom-up mechanism for suggesting different hypotheses, built 
through associative learning mechanisms that encode certain human priors by pretraining on human language.
From an engineering standpoint, they serve as a data-driven proposal distribution over hypotheses $h$ that \emph{might} explain $e_{1:t}$, and where we can down-weight samples that do not fit the data by reweighing to target $p(h|e_{1:t})$, mitigating LLM hallucinations.

To evolve beliefs with each new piece of evidence, we use LLM-augmented Sequential Monte Carlo~\cite{lew2023sequential,zhao2024twistedsmc}, specifically LLM-SMC-S~\cite{Piriyakulkij2024DoingEA} (\cref{fig:smc}).
This maintains $K$ particles $\left\{ h_t^i \right\}_{i=1}^K$ representing candidate hypotheses after observing $t$ examples, $e_{1:t}$.
A prompt implements a bottom-up proposal distribution $q\left( h_{t+1}\mid e_{1:t+1},\left\{ h_t^i \right\}_{i=1}^K \right)$ which
generates a new set of particles $\left\{ h_{t+1}^i \right\}_{i=1}^K$, given the new evidence.
Departing from standard SMC, we propose new particles given a global view of the previous posterior, which means all previous hypotheses are available in-context, and draw $s$ proposals where $s\leq K$ (\Cref{sec:smcappendix}).

A bottom-up associative learner is not the only way of proposing hypotheses, but we think it is close to what happens in humans when drawing fast inferences from sparse data.
Other related cognitive models either curtail the hypothesis space apriori---restricting what can be learned in principle---or demand exorbitant sampling budgets in an effort to cover the vast space of mental programs~\cite{piantadosi2016logical,10.1145/3453483.3454080}.
But an LLM is not the whole story:
Top-down probabilistic reasoning dampens the unpredictability of the language model; allows thinking longer by proposing more hypotheses; and supports a broader range of probabilistic queries, such as asking questions and doing experiments to resolve uncertainty by maximizing information gain.

We intentionally use LLMs,  \emph{not} Large Reasoning Models (LRMs) such as O-series GPT models~\cite{jaech2024openai,guo2025deepseek}, for two reasons.
First, we view human associative learning as closer to the unsupervised pretraining of an LLM than it is to the heavily-supervised post-training of an LRM.
Second, probabilistic reasoning requires generating many plausible inferences, not outputting a single reward-maximizing answer, as LRMs are post-trained to do.

\section{Mental Algorithms from Sequential Observations}

\subsection{List functions}

If humans can infer mental programs, then they should be able to learn new algorithms from examples.
Many 
studies investigate this~\cite{piantadosi2016computational,lake2020people,zhao2024model,hocquette2025humansteachmachinescode,hofstadter1994copycat}, but recently Rule et al.~\cite{rule2024symbolic} substantially increased the behavioral and modeling challenge by testing humans on 250 different algorithms, each learnable from a sequence of examples (\cref{fig:domain}; algorithms 1-100 are easier to model, 101-250 are more challenging).
This benchmark poses a modeling challenge because of the massive combinatorial search space of possible algorithms.
To address this search problem, Rule et al.~\cite{rule2024symbolic} design a custom programming language equipped with high-level search moves (termed \textbf{HL}), 
searching through up to 500k programs for each new input-output to find programs that explain the data.
Plausibly, humans consider far fewer hypotheses---
yet still learn these algorithms.

\begin{figure}[t]
\centering
\includegraphics[alt={Scatter plot of model accuracy versus human accuracy, for our model and Hacker-like model.},width=\textwidth]{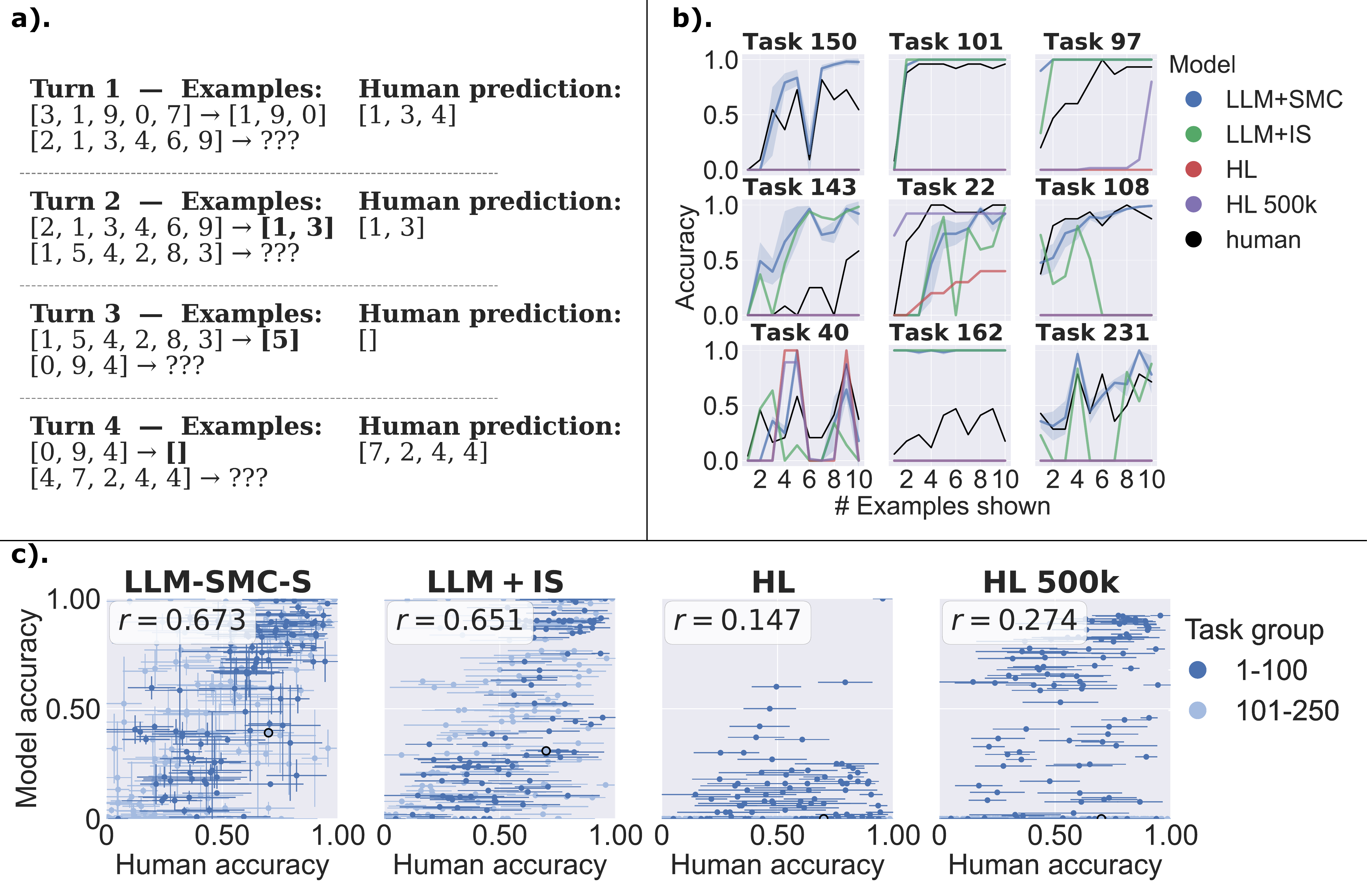} 
\caption{\textbf{a).} An example list function task: the participant is iteratively given examples and asked for predictions. This task -- 150 -- is visualized in panels b). (labeled) and c). (black outline). \textbf{b).} Posterior predictive curves for 4 models, and average accuracy for participants, on 8 randomly drawn tasks, across the 10 examples. Note that accuracy tends to increase as participants / models are shown more examples. \textbf{c).} Scatter plot of model accuracy and human accuracy for the 4 models in panel b).
Model mean accuracy on the List Functions domain across examples versus mean human accuracy across participants and examples ($s=5$ except for HL 500k, where $s=500,000$).}
\label{fig:model_human_acc_scatter}
\end{figure}

We test our model's ability to learn these algorithms while proposing (searching) far fewer hypotheses, and also test our model's ability to capture trial-by-trial dynamics of sequential inference.
To study our ability to predict which algorithms are easier or harder to learn, \Cref{fig:model_human_acc_scatter}c plots human vs. model accuracy on 250 algorithms averaged across trials.
At a search budget of just 5 proposals, our model fits the human data far better than HL given 500k proposals.
This suggests a bottom-up proposal process could explain the search efficiency of human learners:
With a neural proposal distribution, just a few samples suffice to predict average human accuracy.
Modeling the sequence of examples proves important:
Switching from Sequential Monte Carlo to Importance Sampling---which processes all examples at once---degrades model fit (\cref{fig:model_human_acc_scatter}c, LLM+IS).
\Cref{fig:model_human_acc_scatter}b illustrates trial-by-trial accuracy for 8 randomly selected algorithms.
Our model does not capture every detail of these learning curves, but
for 70/100 algorithms, it matches these curves best (under MSE), with the remaining 30/100 roughly equally split between a best fit to HL, and a best fit to IS.

The dataset of Rule et al. served as a significant challenge to both LLMs and conventional symbolic methods.
Humans can learn these algorithms, but it took years of engineering to build a similarly performant model.
Even then, prior work~\cite{rule2024symbolic} confined itself to the 100 easier algorithms, and expended search effort far exceeding what humans plausibly perform.
LLMs alone neither solve these problems nor fit human data, but do so when upgraded with LLM-guided sequential Bayesian updates.

\subsection{Number concepts}

Many human concepts are \emph{categories}, rather than algorithmic functions, and are learned from only positive examples, such telling a child that an animal is `cat', but not saying it isn't a giraffe.
Here we study sequential learning of number categories, such as `numbers ending in 3' or `square numbers bigger than 20', following~\cite{THAKER201710,tenenbaum1999bayesian,ellis2023humanlike}.
When learning such concepts from small amounts of sequential data, humans show ordering effects such as \emph{anchoring} or \emph{garden-pathing}:
A category that seems likely in earlier examples will dominate later inferences, even if invalidated by later data~\cite{macdonald1994probabilistic}.
For example, given the positive examples \emph{30, 31, 33}, humans reliably infer a category such as \emph{numbers around 30}, even when later data suggests multiples of 3, such as \emph{24, 21, 36, 39}.
Ordering the number \emph{31} late in the sequence, such as \emph{30, 33, 24, 21, 36, \underline{31}, 39}, has the opposite effect:
Humans anchor to \emph{multiples of three}, and ignore the non-conforming number 31.

\begin{figure}[t]
\centering
\includegraphics[alt={Number game.},width=0.32\textwidth]{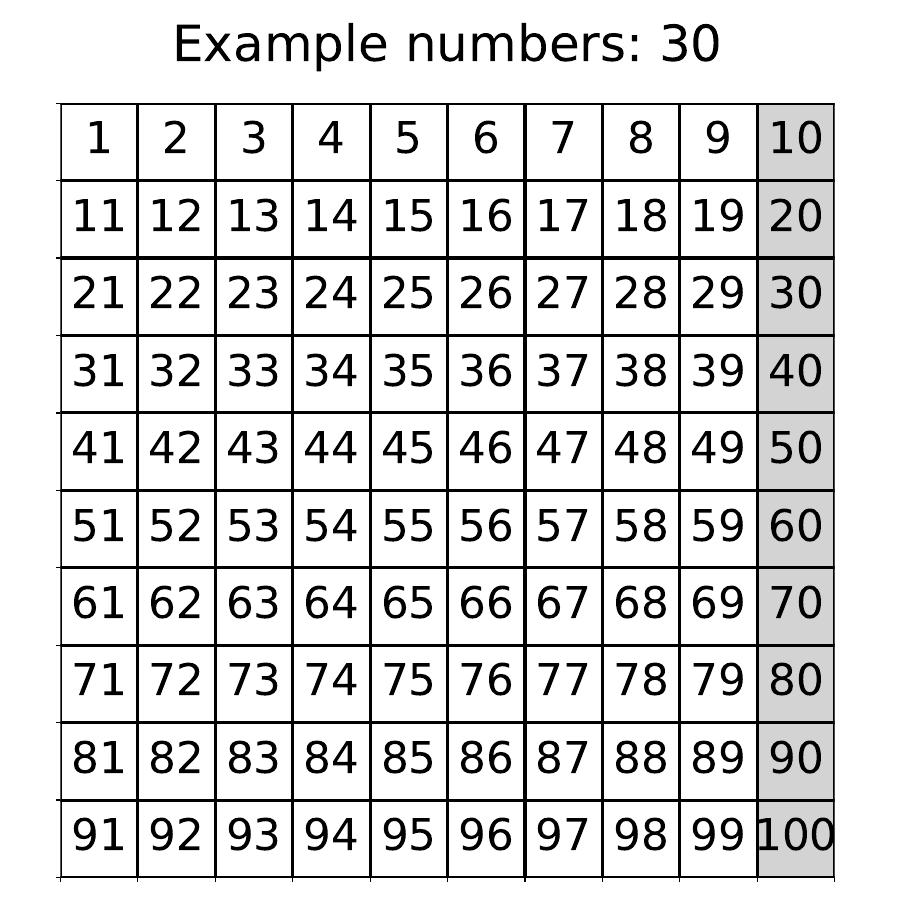} 
\includegraphics[alt={Number game.},width=0.32\textwidth]{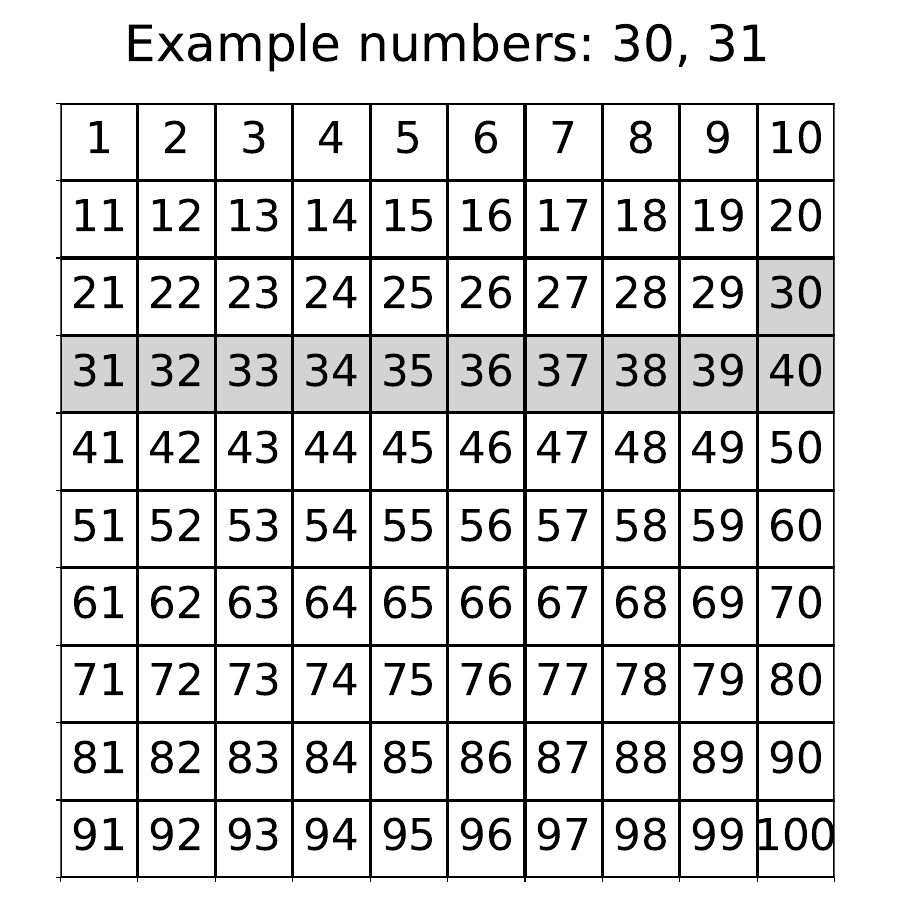} 
\includegraphics[alt={Number game.},width=0.32\textwidth]{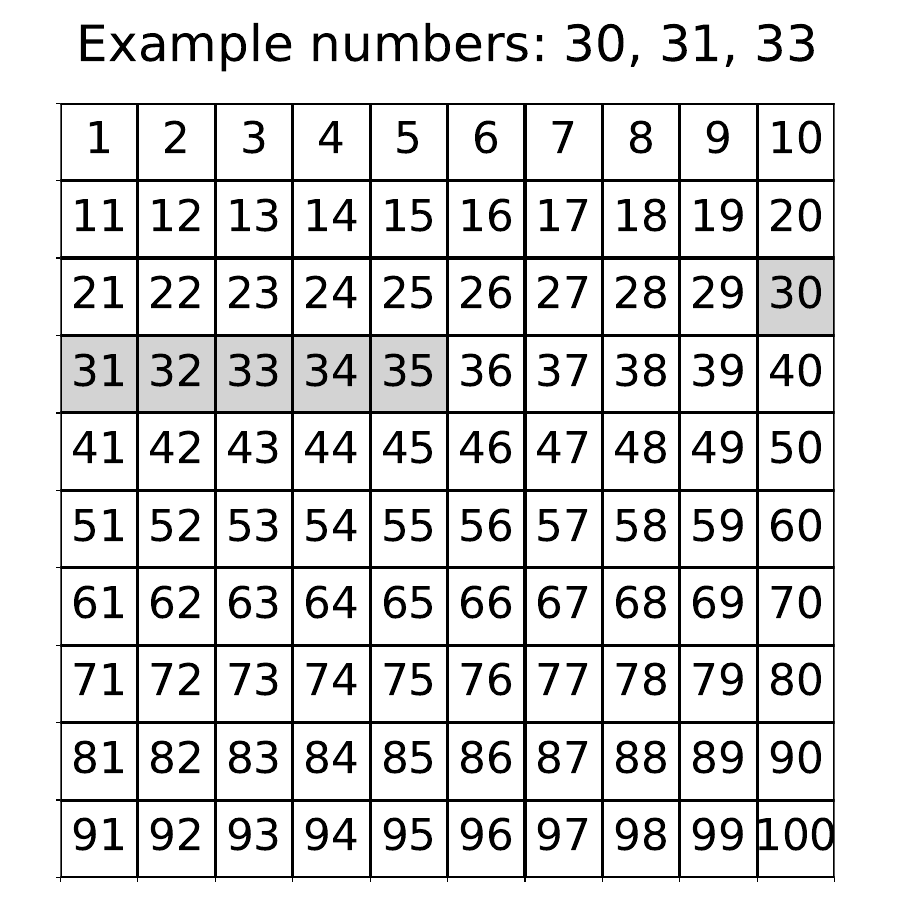} 
\caption{Demonstration of the number game task. Participants are sequentially shown a new example of an unknown number concept and asked to predict what other numbers 1-100 are in the concept (gray).}
\label{fig:number_game_dem}
\end{figure}

\begin{figure}[t]
\centering
\includegraphics[alt={Number game results (distractor).},width=0.7\textwidth]{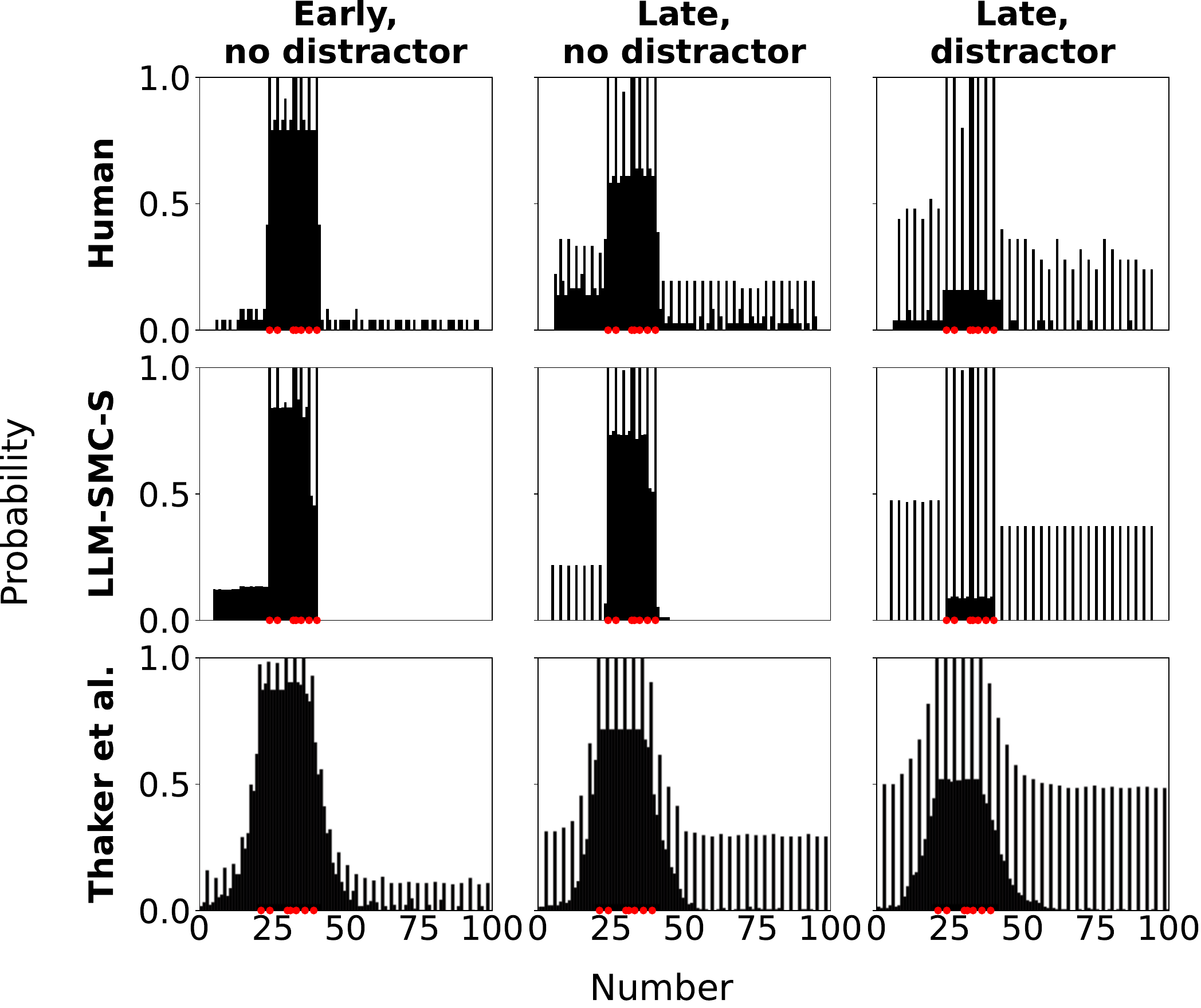}
\caption{Human data, our model predictions, and predictions from Thaker et al.~\citep{THAKER201710} across different orderings (early/late) and distractor conditions. 
In the early condition, learners observe the sequence \emph{30, \underline{31}, 33, 24, 21, 36, 39}.
In the late condition, learners observe \emph{30, 33, 24, 21, 36, \underline{31}, 39}.
In the distractor condition, human participants perform a distractor task (memorizing three distractor numbers), while both computational models have their particle counts reduced (from 70 to 20 for LLM-SMC-S and from 130 to 70 for Thaker et al.).
The early, distractor condition is omitted for clarity (results are nearly indistinguishable from the early, no distractor case). }
\label{fig:number_game_distractor}
\end{figure}

Thaker et al.~\cite{THAKER201710}  study human number-category anchoring (\Cref{fig:number_game_dem}), which we computationally model 
(\Cref{fig:number_game_distractor}).
Our model's sequential inference successfully reproduces the ordering effects seem in humans, and surprisingly, fits the human data better at smaller compute budget than the custom model in Thaker et al.
Our model also reproduces attentional effects.
Thaker et al. induce greater cognitive load in participants by having them perform a distractor task alongside the number game, which causes stronger anchoring.
We model this cognitive load by reducing our number of particles, which replicates the effect.

\section{Resolving Uncertainty by Doing Experiments and Asking Questions}

Humans can procure new data to aid learning by asking questions or trying out experiments in the real world, such as eating a new kind of berry to tell if it makes us sick.
But it costs something to acquire new learning data, so
humans need to decide whether to incur the cost of doing an experiment or asking a question to resolve uncertainty.
Building on our sequential inference setup, we treat humans as considering different experiments $\experiment$, and pick the experiment which maximizes expected information gain, under their particle-based approximate probabilistic beliefs: 
\begin{align}
\experiment^* &= \argmax_{\experiment} \expect_{p(e|\experiment, e_{1:t})} 
\KL\left(p\left(h\mid e_{1:t}, e\right) \;\;\middle\|\;\; p\left(h\mid e_{1:t}\right)\right) 
\label{eq:infogain}
\end{align}
Exact expected information gain computation is intractable.
We make a particle-based approximation by treating the hypothesis space as the set of unique current particles to approximate the distributions $p(h\mid e_{1:t})$, $p(e\mid\experiment,e_{1:t})$, and $p(h\mid e_{1:t},e)$.
Maximizing over all possible experiments is also intractable---infinitely many exist---so instead, an LLM proposes finitely many experiments, which are scored under \cref{eq:infogain}.

\begin{figure}[!b]
    \centering
    \begin{subfigure}[b]{\textwidth}
        \caption{}
        \centering
        \includegraphics[width=0.7\textwidth]{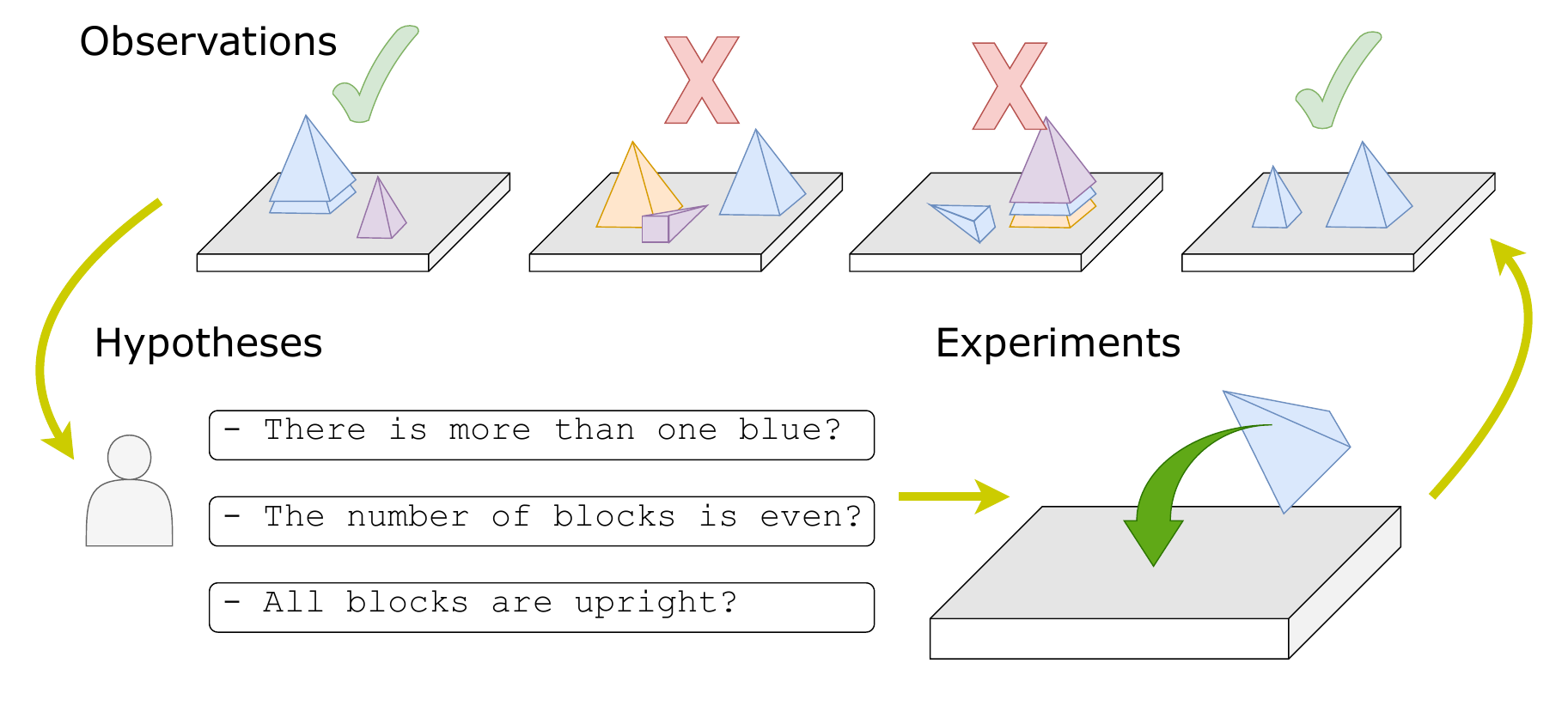}
        \vspace{-2em}
        \label{fig:zendo_gameply}
    \end{subfigure}
    \vskip\baselineskip
    \begin{subfigure}[b]{\textwidth}
        \caption{}
        \centering
        \includegraphics[width=\textwidth]{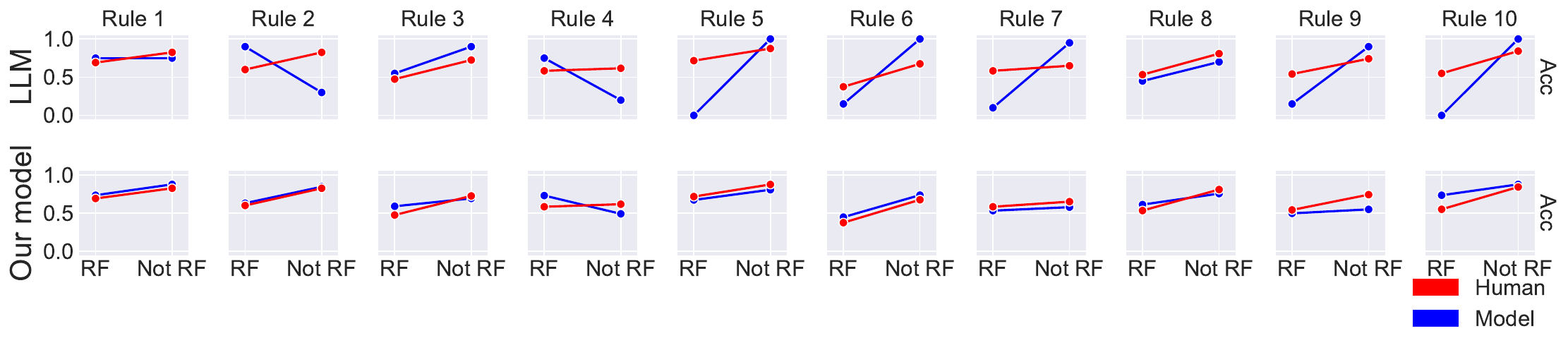}
        \vspace{-3em}
        \label{fig:sub3}
    \end{subfigure}
    \vskip\baselineskip

    \begin{subfigure}[b]{0.50\textwidth}
        \caption{}
        \centering
        \includegraphics[width=\textwidth]{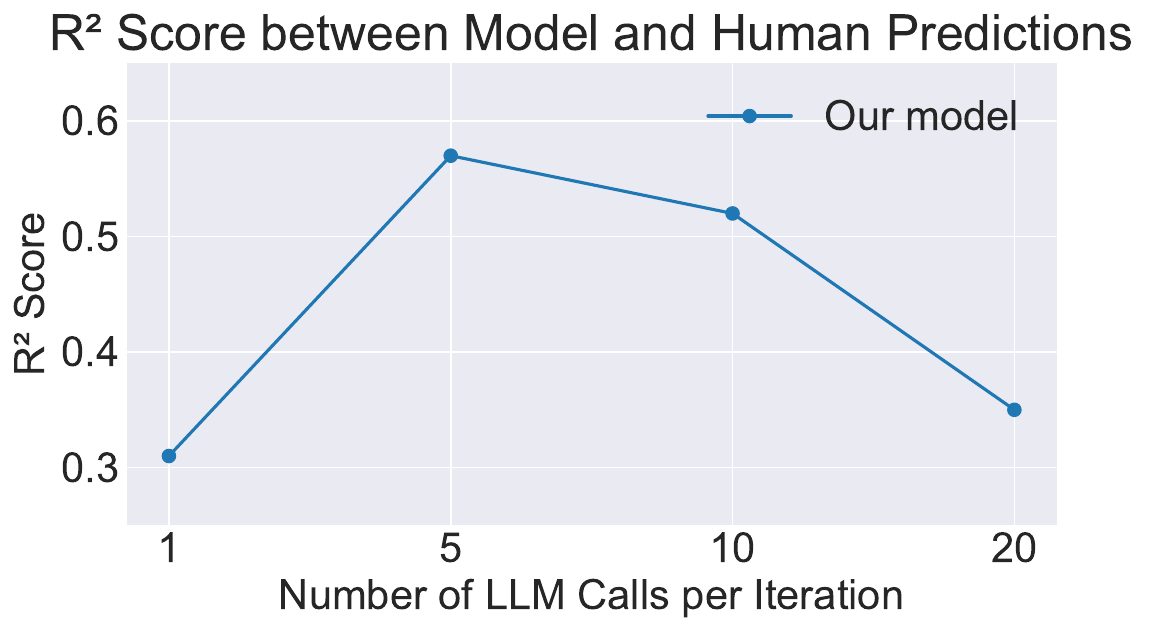}
        \label{fig:sub3}
    \end{subfigure}
    \hfill
    \begin{subfigure}[b]{0.49\textwidth}
        \caption{}
        \centering
        \raisebox{2.5cm}{
            \begin{tabular}{c@{\hspace{5pt}}c}
            \toprule
            Method & LogL$\uparrow$   \\
            \midrule
            Bramley et al. 2018~\cite{bramley2018grounding}          & $-1539$         \\
            
            
            Important Sampling + Refinement 
            & $-3499.76$  \\
            
            
            Importance Sampling               & $-1660.90$        \\
            
            LLM-SMC-S             & $\bm{-1478.82}$        \\
            \bottomrule
            \end{tabular}
        }
        \label{fig:sub4}
    \end{subfigure}
    \vspace{-2em}
    \caption{(a) Zendo gameplay 
    (b) Human accuracy on examples that do not belong to the category (not rule following, `Not RF') is higher than accuracy on in-category examples (rule following, `RF'). Our model reproduces this, an LLM on its own does not. (c) Bounded rationality: Human-model $R^2$ has a U-shaped relationship to compute budget.
    Humans are boundedly rational, and increasing compute eventually degrades model fit.
    (d) Log likelihood of the human data under different models.
    }
    \label{fig:zendo}
\end{figure}

We first investigate this model by comparing its behavior to humans playing the game Zendo, which resembles classic `blicket' studies in developmental psychology~\cite{gopnik2000detecting, COOK2011341, zhang2021acre}, but adds active experimentation (\Cref{fig:zendo}a).
In Zendo, players infer a hidden binary category by building constructions from colored shapes, and then receiving feedback on if their construction belongs to the hidden category 
Each construction is an experiment $\experiment$.
We take human data from~\cite{bramley2018grounding}, where after 7 rounds of experimentation, participants make 8 predictions on holdout test constructions.
Our model mimics a human participant by alternating between experimentation and inference, and finally testing on the the same holdout constructions.
The resulting model captures fine-grained structure in human error patterns (\cref{fig:zendo}b).
As before, we get insight into boundedly-rational human behavior by modulating the inference-time budget (\cref{fig:zendo}c), and despite minimal domain-specific engineering, our model fits the human data better than a custom Bayesian learner designed specifically for this dataset (\cref{fig:zendo}d).
We find therefore that the LLM-guided Bayesian learner is surprisingly versatile:
Rational probabilistic reasoning tied to LLM backends support both concept learning and active experimentation,
giving an induction-inquiry cycle that predicts human behavior better than LLMs or Bayes on their own.

\paragraph{Beyond Code: Asking Questions.}
In social contexts, the analog of an experiment is question-asking.
Because our model operates over natural language, 
we can also use it to generate informative questions.
We study this in a web shopping task where the model serves as a shopping assistant, and asks natural-language questions that optimize information gain (\cref{eq:infogain}).
As a proof-of-concept, we assume a finite hypothesis space of products a customer can purchase, and perform exact Bayesian belief updates.
An experiment $\xi$ is a natural language question, an example $e$ is a question-answer pair, and hypotheses $h$ are different products (such as different brands of shampoo).

The agreement of a product $h$ and a question-answer pair $e$ is not generally expressible as a Python program.
Therefore the likelihood $p(e|h)$ simply queries a language model, rather than generate code.
Conceptually our argument is that \emph{neither} natural language nor high-level programming languages can singlehandedly capture the biases and expressiveness of human mental programs, so we should expect that, for some domains, its suffices to use only language (or only code). 

\begin{figure*}[!h]
\centering
\includegraphics[width=0.48\textwidth]{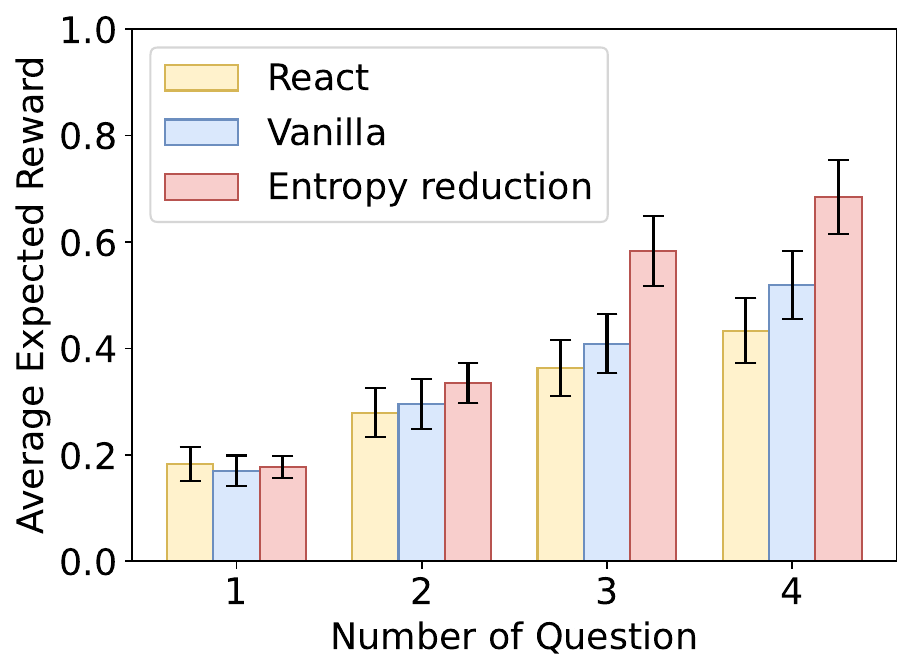}
\includegraphics[width=0.48\textwidth]{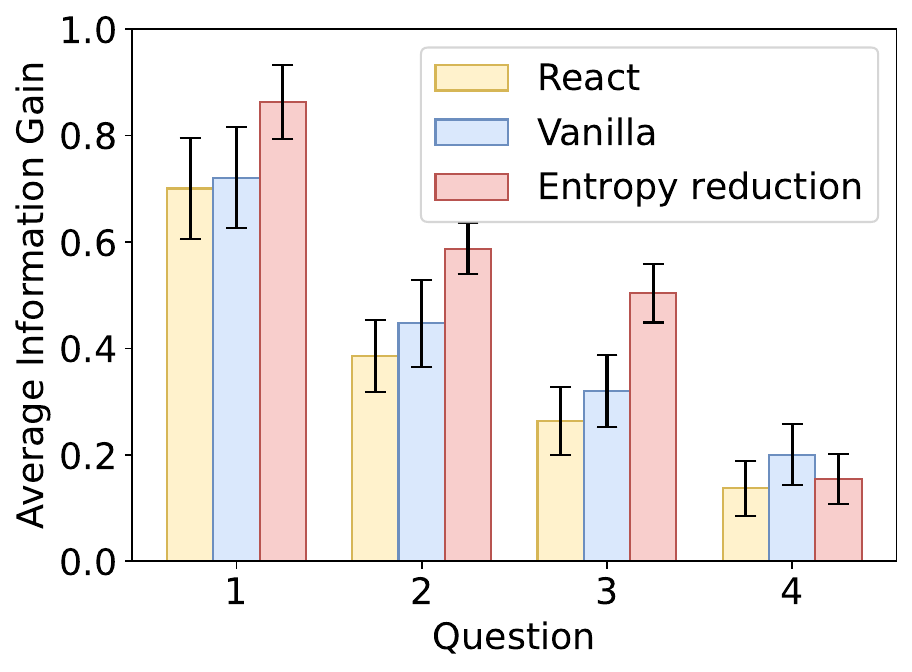}
\vspace{-1em}
\caption{(Left) Average expected binary reward at increasing number of number of questions. (Right) Average information gain at each question.}
\label{fig:webshop}
\end{figure*}




\Cref{fig:webshop} compares our model with basic LLM prompts and also ReAct, which prompts an LLM to think before it acts. Our model chooses more informative questions, as defined by information gain, 
giving a higher chance of discovering  the customer's preferred product. 
Although we mainly focus on modeling human data, we believe that these methods  can also impact how AI systems are built.

\section{Discussion}

How people learn new categories, laws, and abstract concepts from the sparse streaming data of experience is a difficult open question.
Candidate computational models must be simultaneously flexible and efficient---which are generally in tension.
To make progress on that question, our models make mechanistic commitments about the underlying mental representation---code and language---and the underlying mental algorithms, which use neurally-guided sequential inference for tractable reasoning over open-ended hypothesis spaces.
Across a range of inductive reasoning problems, the resulting model fits human data better than specialized models designed for each individual behavioral experiment, and further allow induction to alternate with inquiry, explaining how online learning and acting can cooperate together.
Our results suggest a language of thought that lies between logic and language, and also suggest that we are not far from a unified computational account of human induction and inquiry that could explain how humans think flexibly across the endless range of situations in which these cognitive faculties can be brought to bear.

\paragraph{Cognitive Implications.}

Our models assume a solution space whose hypotheses are at least definable in clear English language.
Yet many human concepts are famously tricky to formalize, such as the meaning of `chair'~\cite{wittgenstein1953philosophical} or even `dog walking'~\cite{quinn2016activeobjectlocalizationvisual}.
It remains open whether symbolic hypotheses are the right representation for such categories, but our view is that a symbolic language of thought remains the best account of inductive reasoning from small data.
We use a specific language of thought termed \emph{mental programs}, which combines natural language and code.
Code can represent what would be hard to precisely express language, such as detailed physical properties, the shape of a hand-drawn character~\cite{lake2015human}, or the precise rules of a board game.
Language can represent fuzzier, higher-level propositions.
Current AI heavily uses language as a knowledge representation, and achieves unprecedented coverage as a result:
LLMs can converse about virtually every human topic, even if they do not always make sense.
Classic Bayesian models using symbolic programs have succeeded at  modeling humans within narrow domains by tailoring the representation to the problem domain~\cite{goodman2008rational, amalric2017language, ellis2022synthesizing, bramley2018grounding, erdogan2015sensory, sablemeyer2017geom, tian2020learning, saad2019bayesian}.
Our work suggests that Bayesian priors should operate over in language-like representations, but that the grounding between hypothesis and data---the likelihood function---is a better fit for programs.
Potentially, a future unified representation could serve both roles, which we view as an important open direction.

Bayesian cognitive models are known both for their predictive power and their computational intractability, and have been criticized as lacking an account of their biological implementation.
Our work is not alone in trying to address these issues with Bayesian models:
Metalearning offers another tractable neural instantiation of Bayes, where a neural network approximates the Bayesian predictive distribution~\cite{lake2023human,xieexplanation,grant2018recasting} in a single forward pass, either via in-context learning or via MAML-style~\cite{finn2017model} weight updates.
Unlike our work, such models do not reason over latent discrete representations.
This is complementary to our models:
We construct explicit verbalizable hypotheses, do not require training new neural networks, and can trade more inference-time compute for better predictions.
At the same time, relative to metalearning, our approach has important limitations:
It cannot learn what pretrained models do not already understand, and is unlikely to be a good account of fast, non-verbalizable inference.
Roughly, we think of our models as a System 2 way of using neural networks for approximate reasoning, while metalearning is best thought of as a fast System 1 process. 
Humans likely use both strategies when thinking probabilistically.

\paragraph{Neural networks and Bayes.}
LLMs are probabilistic models trained on masses of human data---yet, in isolation, they do not reproduce human behavior in the tasks considered here.
Why is that?
Fundamentally, probabilistic inference requires reasoning about uncertainty.
Reasoning requires expending variable compute, to think longer on harder problems.
Handling uncertainty further constrains the model to the laws of probability.
The newest LLMs and their successors, LRMs, are increasingly trained to reason via reinforcement learning (using chain-of-thought~\cite{deepseekai2025deepseekr1incentivizingreasoningcapability}), but to date such training focuses on problem-solving, not probabilistic inference.
Chain-of-thought may also not be sufficiently constrained to ensure sound convergence, unlike the classic Monte Carlo algorithms that we and others build on~\cite{zhao2024twistedsmc}.
It remains open however whether future LLMs could implicitly learn to mimic the reasoning patterns of sound inference algorithms. 



\paragraph{Outlook.} Neural networks trained on massive human data can reach human-level performance in many areas, and are increasingly recognized as valuable models of human thinking~\cite{binz2025foundation}.
This leaves open however the scientific questions how human minds grow into their adult competencies without such data, and leaves open how these models could surpass what they see in training data.
In our view, a missing ingredient is the rational analysis 
of the computational problems and their optimal solutions, which constrains and informs the training and use of these models.
Here we develop such a rational analysis for induction and inquiry, finding it complements rather than competes with modern LLMs.
Our particular rational analysis could also help in areas of AI such as automated scientific discovery, forcasting and prediction, and embodied (or digital) agents.
Within cognitive science, our broader paradigm could apply more generally to social reasoning, metareasoning, the growth and maintenance of intuitive theories~\cite{carey1985conceptual}, and many other areas where people efficiently think and learn.




\bibliographystyle{unsrt}
{\small \bibliography{CogSci_Template}}

\clearpage
\appendix
\section{Appendix}\label{sec:appendix}

\subsection{LLM-SMC-S details}\label{sec:smcappendix}

We describe the pseudocode for LLM-SMC-S in \Cref{fig:smcs_pseudocode}.

\begin{algorithm*}[!b]
\caption{LLM-SMC-S algorithm}
\begin{algorithmic}
\State Let $e_1=(x_1, y_1)$ be the first data point we observe
\State $h_1^{(1)}, ..., h_1^{(n)} \sim q(h|x_1, y_1)$
\State $w_1^{(i)} \gets \frac{p(x_1, y_1, h_1^{(i)})}{q(h|x_1, y_1)}$ for $1 \leq i \leq n$ \Comment{Reweighting}
\State $H_1 \gets \textit{Resampling}(H_1, W_1)$ \Comment{Resampling}

\For{$t = 2, ..., T$}
\State The active learning algorithm gives $e_t=(x_t, y_t)$
\State $h_t^{(1)}, ..., h_t^{(n)} \sim q(h|H_{t-1}, x_{1:t}, y_{1:t})$ \Comment{Rejuvenating}
\State $A(h_t^{(i)}, H_{t-1}, W_{t-1}) = \frac{1}{n} \sum_{j=1}^n w_{t-1}^{(j)} \frac{p(h_t^{(i)}|x_{1:t}, y_{1:t})r(h_{t-1}^{(j)}|h_t^{(i)}, x_{1:t}, y_{1:t})}{p(h_t^{(j)}|x_{1:t-1}, y_{1:t-1})}$
\State $w_t^{(i)} \gets \frac{A(h_t^{(i)}, H_{t-1}, W_{t-1})}{q(h_t^{(i)}|H_{t-1}, x_{1:t}, y_{1:t})}$ for $1 \leq i \leq n$ \Comment{Reweighting}
\State $H_t \gets \textit{Resampling}(H_t, W_t)$ \Comment{Resampling}
\EndFor
\end{algorithmic}
\label{fig:smcs_pseudocode}
\end{algorithm*}

For all domains in this paper, we define the forward kernel $q$ using the following format:
\begin{align}
q(h|H, e_{1:t}) &\propto \bone[h \in (H \cup B(e_{1:t}, H))]
\end{align}
where $B$ can be any deterministic, LLM-based functions that output a set of hypotheses.

We use GPT-4 as our main LLM for our experiment.

\subsection{Instantiations of the probabilistic model in each domain}

Here, we describe how instantiate the prior and likelihood in our probabilistic model described in \Cref{eq:beliefupdate} for each domain:

\paragraph{List functions and Number concepts.}

The prior is defined to be $p(h) \propto \frac{1}{|h|}$ for list functions and $p(h) \propto e^{-word\_count(h)}$ for number concepts.

The likelihood is defined as follows:
\begin{equation}
    p(e_{1:T}\mid h) =
\prod_{t\leq T} (1-\theta)\frac{\mathds{1} [y_t = h(x_t)]}{T} + 
\theta \frac{\mathds{1} [y_t \neq h(x_t)]}{T}
\end{equation}
where $e = (x,y)$ and $\theta$ is the probability that an example is mislabeled. We set $\theta=\frac{1}{100}$, and after proposing all hypotheses, we fit $\theta$ to human data using $k$-fold cross-validation, with $k=10$.

\paragraph{Zendo.}

For this domain, we extend the model to include false-positives and false-negatives random variables. 
Our likelihood model is then defined as:
\begin{align}
    p(y=1|x, h, \epsilon, \delta) = \; \left[
    \begin{array}{cc}
         \delta &\text{ if } {h}(x) = 1\\
         1-\epsilon &\text{ if } h(x)=0
    \end{array}\right]
\end{align}

Both $p(\delta)$ and $p(\epsilon)$ are normal distributions, truncated to remain within the range [0.5, 1], with $\mu=0.7,\sigma=0.1$ and $\mu=0.9,\sigma=0.01$ respectively. 

For $p(h)$, we define $p(h) \propto (\frac{1}{word\_count(h)})^2$.

\end{document}